\documentclass[letterpaper, 10 pt, conference]{ieeeconf}  

\IEEEoverridecommandlockouts                              

\usepackage[usenames, dvipsnames]{color}
\usepackage{listings}
\usepackage{xcolor}
\usepackage{url}
\usepackage{float}
\usepackage{amsmath,amssymb,amsthm}
\usepackage{mathtools} 
\usepackage[english]{babel}

\usepackage{epsfig}
\usepackage{epstopdf}
\usepackage{subcaption}
\usepackage{booktabs}
\usepackage{multirow}
\usepackage{siunitx}
\usepackage{graphicx}

\usepackage[export]{adjustbox}
\usepackage[boxruled,vlined,linesnumbered]{algorithm2e}

\usepackage{hyperref}
\usepackage[autostyle, english = american]{csquotes}
\MakeOuterQuote{"}

\hypersetup{
    colorlinks=true,
    linkcolor=blue,
    }

\title{\LARGE \bf
Non-Trivial Query Sampling For Efficient Learning To Plan
}

\author{Sagar Suhas Joshi$^{1}$~ Panagiotis Tsiotras$^{2}$
	\thanks{$^{1,2}$ Institute for Robotics and Intelligent Machines, Georgia Institute of Technology, USA.
		Email: {\small sagarsjoshi94@gmail.com}}
}

\begin{document}

\maketitle
\thispagestyle{empty}
\pagestyle{empty}

\begin{abstract}
In recent years, learning-based approaches have revolutionized motion planning. 
The data generation process for these methods involves caching a large number of high quality paths for different queries (start, goal pairs) in various environments. 
Conventionally, a uniform random strategy is used for sampling these queries. 
However, this leads to inclusion of "trivial paths" in the dataset (e.g.,, straight line paths in case of length-optimal planning), which can be solved efficiently if the planner has access to a steering function.  
This work proposes a "non-trivial" query sampling procedure to add more complex paths in the dataset.
Numerical experiments show that a higher success rate can be attained for neural planners trained on such a non-trivial dataset.
\end{abstract}
\section{INTRODUCTION}
Motion planning, a core problem in artificial intelligence and robotics, is one of finding a collision free, low cost path connecting a start and goal state in a search-space. 
Popular discrete-space planners such as A*~\cite{hart1968formal} and LPA*~\cite{koenig2004lifelong} conduct a prioritized search using heuristics and guarantee resolution-optimal paths. 
On the other hand, single-query sampling-based motion planning (SBMP) algorithms, such as RRT~\cite{lavalle2001randomized}, solve this problem by constructing a connectivity graph online using a set of probing samples.
The RRT algorithm is \textit{probabilistically complete}, while it's \textit{asymptotically optimal} variants such as RRT*~\cite{karaman2011sampling}, RRT$^{\#}$~\cite{arslan2013use}, BIT*~\cite{gammell2015batch}, FMT*~\cite{janson2015fast} converge to the optimal solution almost surely, as the number to samples tends to infinity.  
However, these algorithms may suffer from a slow convergence rate, especially in higher dimensional settings. 
In order to address this issue, several techniques that leverage heuristics and collision information have been suggested to improve the performance of these planners. 
These include \cite{phillips2004guided}, \cite{persson2014sampling}, \cite{akgun2011sampling}, \cite{rodriguez2006obstacle}, \cite{urmson2003approaches} among others.
The recently proposed Informed Set~\cite{gammell2018informed}, \cite{joshi2020timeinformed}, \cite{mandalika2021guided} and  Relevant Region~\cite{joshi2019relevant}, \cite{arslan2015dynamic} family of algorithms utilize current solution information and heuristics to focus the search onto a subset of the search-space, while still maintaining the theoretical guarantees of asymptotic optimality.  

Although the above methods can improve the performance of motion planning algorithms, they require handcrafted heuristics for efficacy. 
Also, these methods do not leverage prior experience or data gathered from expert demonstrations. 
Deep learning based approaches for motion planning address these two limitations by generating a dataset of high quality paths in various environments. 
In the offline phase, this dataset is used to train a deep neural network (DNN) model to predict quantities of interest, such as cost-to-go~\cite{Chen2020Learning}, next point along the optimal path~\cite{qureshi2020motion}, or a sampling distribution~\cite{ichter2018learning}.
The DNN model can then be used online to focus search and dramatically increase the efficiency of planning algorithms. 

\begin{figure}
    \centering
    \includegraphics[width=0.58\columnwidth]{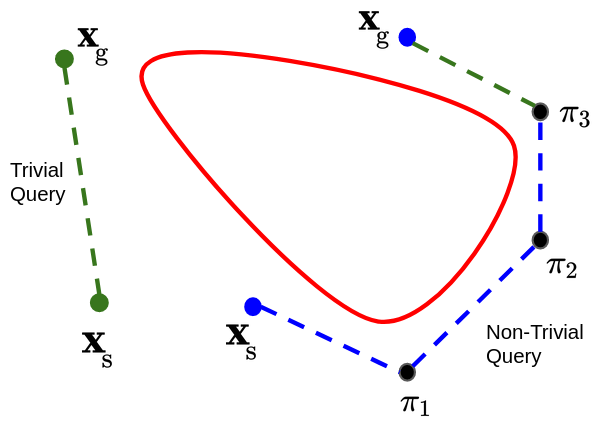}
    \caption{Schematic for the proposed data generation method. Instead of \textcolor{green}{trivial queries} that can be solved by a greedy connection, more \textcolor{blue}{non-trivial queries} are added. In addition, the data pruning procedure filters out the trivial part $(\pi_3,\textbf{x}_\mathrm{g})$ of the path.}
    \label{fig:non_trivial_schematic}
\end{figure}

The data generation process for DNN-based motion planning methods involves sampling and solving a set of queries (start, goal pairs) in a given environment.
For many applications, the map/environment in which the robot needs to operate may be fixed or given apriori.
However, the query sampling distribution can be modified in order to extract more informative paths beneficial to the learning process.
Conventionally, uniform random sampling is used to generate the start and goal states. 
Many queries in this uniformly sampled dataset can be solved by greedily connecting the start and goal state using a steering function (if available).
Such a steering function provides the optimal path between any two states after relaxing the collision constraint.
This work proposes adding more "non-trivial" queries to the dataset, which cannot be solved by a simple greedy connection.
The efficiency of the neural planners can be boosted by training deep models on this dataset comprising of relatively more complex paths. 
This is demonstrated by creating datasets with different degrees of non-triviality and benchmarking the performance of the neural planner on various robotic planning tasks.

\section{RELATED WORK}
Exciting progress has been made at the intersection of learning and motion planning in recent years.
To improve the performance of discrete space planners, methods that leverage  reinforcement learning \cite{zucker2008adaptive} and imitation learning \cite{bhardwaj2017learning}, \cite{choudhury2018data} have been proposed.  
Neural A* \cite{yonetani2021path} presents a data-driven approach that reformulates conventional A* as a differentiable module.
Huh et al \cite{huh2020learning} present a higher-order function network that can predict cost-to-go values to guide the A* search. 
For sampling-based planning, techniques such as \cite{chamzas2019using} and \cite{lai2020bayesian} learn a local sampling strategy for global planning.
Zhang et al \cite{zhang2018learning} present a deep learning based rejection sampling scheme for SBMP.
Ichter et al~\cite{ichter2018learning} train a conditional variational autoencoder (CVAE) model to learn sampling distributions and generate samples along the solution paths. 
NEXT~\cite{Chen2020Learning} learns a local sampling policy and cost-to-go function using "meta self-improving learning".
Kuo et al~\cite{kuo2018deep} input the planner state and local environment information into a deep sequential model to guide search during planning.
Approaches such as \cite{ichter2020learned}, \cite{Kumar2019LEGOLE} focus on identifying critical or bottleneck states in an environment to generate a sparse graph while planning. 
While the above techniques may differ in terms of their model architectures or outputs, they do not tackle the problem of improving the data-generation process via query sampling for increasing the efficacy of neural planners.

\IncMargin{.5em}
\begin{algorithm}[t]
	\caption{Neural Planner}
	\label{alg:neuralplanner}
	\SetKwFunction{NeuralPlanner}{}
	\SetKwProg{Fn}{NeuralPlanner}{:}{}
	\Fn{\NeuralPlanner{ $ \textbf{x}_\mathrm{s},\textbf{x}_\mathrm{g},\mathcal{X}_\mathrm{obs} $ }}
	{
	    $\pi \leftarrow \{\textbf{x}_\mathrm{s}\} $\;
	    \For{$i=1:N_\mathrm{plan}$}
	    {
	        \eIf{$\mathsf{steerTo}(\pi_\mathrm{end},\textbf{x}_\mathrm{g},\mathcal{X}_\mathrm{obs})$}
	        {
	            $\pi \leftarrow \pi \cup \{\textbf{x}_\mathrm{g}\} $\;
	            $\mathsf{break}$\;
	        }
	        {
	            $\textbf{x}_\mathrm{new} \leftarrow \mathsf{PNet}(\pi_\mathrm{end},\textbf{x}_\mathrm{g}) $\;
	            $\pi \leftarrow \pi \cup \{\textbf{x}_\mathrm{new}\} $\;
	        }
	    }
	    \If{$\textbf{x}_\mathrm{g} \notin \pi$}
	    {
	        \KwRet $\emptyset$ \;
	    }
	    \eIf{$\mathsf{Feasible}(\pi)$}
	    {
	        \KwRet $\pi$\;
	    }
	    {
	        \KwRet $\mathsf{Replan}(\pi)$\;
	    }
	}
	
\end{algorithm}
\DecMargin{.5em}

In \cite{huh2021learning}, Huh et al extend their previous work to present a query sampling technique for non-holonomic robots.
However, this technique is purely based on the dynamics of the robot, does not utilize obstacle information and is only applicable for car-like robots.
The recently proposed OracleNet \cite{bency2019neural} and Motion Planning Networks (MPNet) algorithm \cite{qureshi2020motion} learn a deep model to recursively predict the next state along the solution path, given a query.
The authors in \cite{qureshi2020motion} use "Active Continual Learning (ACL)" to improve the data-efficiency of the training process. 
Similar to the DAGGER algorithm~\cite{ross2011reduction} the ACL process involves training a MPNet deep model on a set of expert demonstrations for a $N_c >0$ number of initial iterations.
The ACL algorithm then finds the cases where the MPNet planner fails and invokes the expert planner only to solve them.
The solutions generated by the expert planner are stored in a replay buffer to be used during training.
ACL also leverages the "Gradient Episodic Memory (GEM)" technique \cite{lopez2017gradient} to alleviate the problem of catastrophic forgetting during the learning process.
While more data-efficient, the ACL process can be tricky to implement and computationally more expensive, as it involves interleaving the training process with running the neural planner multiple times.  
The performance of ACL models can also be worse than that of batch-offline models in many cases \cite{qureshi2020motion}.
In contrast, this work proposes a modified query sampling procedure for data-generation and trains all models in a batch-offline manner.

\section{PROBLEM DEFINITION}

\subsection{Path Planning Problem}
Let $\mathcal{X} \subset \mathbb{R}^d$ denote the search-space for the planning problem with dimension $d \geq 2$.
Let $\mathcal{X}_\mathrm{obs} \subset \mathcal{X}$ denote the obstacle space and $\mathcal{X}_\mathrm{free}= \mathcal{X}\setminus \mathcal{X}_\mathrm{obs}$ denote the free space.
Let $\mathrm{c}_{\pi}(\textbf{x}_\mathrm{s},\textbf{x}_\mathrm{g})$ denote the cost of moving from a point $\textbf{x}_\mathrm{s} \in \mathcal{X}$ to $\textbf{x}_\mathrm{g} \in \mathcal{X}$ along a path $\pi$.
This path $\pi$ can be represented as an ordered list of length $L\geq2$, $\pi=\{\pi_{1},\pi_{2} \dots \pi_{L} \}$. 
Let $\pi_\mathrm{end}$ denote the last point on the path $\pi$.
Then, the optimal path planning problem is one of finding the minimum cost, feasible path $\pi^*$ connecting a start $\textbf{x}_\mathrm{s}$ and goal $\textbf{x}_\mathrm{g}$ state.
\begin{equation}
\label{eq:optimalPlanningDef}
\begin{aligned}
\pi^*=\arg \min_{\pi \in \Pi}  & \ \mathrm{c}_{\pi}(\textbf{x}_\mathrm{s},\textbf{x}_\mathrm{g}), \\
\text{subject to:} & \ \pi_{1}=\textbf{x}_\mathrm{s}, \pi_\mathrm{end}=\textbf{x}_\mathrm{g} \\
& \ \pi_\mathrm{i} \in \mathcal{X}_\mathrm{free}, ~~~ i = 1,2 \dots L.  
\end{aligned}
\end{equation}
Classical planners discretize $\mathcal{X}_\mathrm{free}$ or build a connectivity graph and then perform a search over this graph to solve the above planning problem (\ref{eq:optimalPlanningDef}).
\subsection{Supervised Learning for Planning}
Learning-based methods use the data gathered from successful plans to train models in the offline phase. In the online phase, this learned model can be used to solve (\ref{eq:optimalPlanningDef}) or assist the classical planners.
The data generation process involves creating an environment and sampling a set of $K_\mathrm{train}>0$ queries or (start, goal) pairs in $\mathcal{X}_\mathrm{free}$. 
Let $\mathcal{Q}=\mathcal{X}_\mathrm{free} \times \mathcal{X}_\mathrm{free}$ denote the "query-space".
A classical planner is then used to solve the path planning problem (\ref{eq:optimalPlanningDef}) for each of the $K_\mathrm{train}$ queries and obtain a good quality solution.
A quantity of interest to be learned as output (such as cost-to-go, next state on the optimal path etc) is extracted from these solution paths. 
The objective of the training process is to learn a function $f_\theta$ (usually a deep neural network), on the dataset $\mathcal{D}$ by minimizing an empirical loss with respect to weight parameters $\theta$
\subsection{Neural Planner}
The MPNet procedure involves learning a planning network $\mathsf{PNet}$, that predicts the next state on the optimal path, given a current state, a goal state and environment information as the input.  
A typical neural planning algorithm, in line with the one described in \cite{qureshi2020motion}, is illustrated in Algorithm~\ref{alg:neuralplanner}.
Given a query $\textbf{x}_\mathrm{s},\textbf{x}_\mathrm{g}$, the path $\pi$ to be returned is initialized with the start-state. 
At each iteration, the neural planner attempts a greedy connection to the goal using the $\mathsf{steerTo}$ function. 
For length-optimal or geometric planning case, the $\mathsf{steerTo}$ connects any two points using a straight line.
For car-like robots, this steering function can use the Dubins curves for dynamically feasible connections \cite{johnson2020dynamically}.
To probe the feasibility of this greedy connection, the $\mathsf{steerTo}$ procedure discretizes the path and collision-checks the points on it.
If the greedy connection $\pi_\mathrm{end}$ to $\textbf{x}_\mathrm{g}$ is valid, the goal-state is appended to the path and the planning loop terminates. 
Else, the learned planning network $\mathsf{PNet}$, is used to predict the next state on the optimal path. 
If the path $\pi$ is infeasible, a neural replanning procedure is performed on this coarse path in an attempt to repair it. 
Please see \cite{qureshi2020motion} for more details about these procedures.

\section{NON-TRIVIAL QUERY SAMPLING}
\IncMargin{.5em}
\begin{algorithm}[t]
	\caption{Data Generation Algorithm}
	\label{alg:dataGen}	
	$\mathcal{D} \leftarrow \emptyset$\;
	$\mathcal{X}, \mathcal{X}_\mathrm{obs} \leftarrow \mathsf{createEnvironment}()$\;
	\For{$j=1:K_\mathrm{train}$}
	{
	    $u_\mathrm{rand} \sim U[0,1]$\;
	    \eIf{$u_\mathrm{rand}<p_\mathrm{nt}$}
	    {
	       $\textbf{x}_\mathrm{s},\textbf{x}_\mathrm{g} \leftarrow \mathsf{nonTrivialQuerySampling}(\mathcal{X}_\mathrm{free})$\;
	    }
	    {
	        $\textbf{x}_\mathrm{s},\textbf{x}_\mathrm{g} \leftarrow \mathsf{uniformSampling}(\mathcal{X}_\mathrm{free})$\;
	    }
	    $\pi \leftarrow \mathsf{solveQuery}(\textbf{x}_\mathrm{s},\textbf{x}_\mathrm{g},\mathcal{X}_\mathrm{obs} )$\;
	    $\mathcal{D} \leftarrow \mathsf{includeData}(\mathcal{D},\pi) $\;
	}
	\KwRet $\mathcal{D}$
\end{algorithm}
\DecMargin{.5em}

\subsection{Non-trivial Queries}
Conventionally, uniform random sampling is used in the data generation process to obtain a query $\textbf{x}_\mathrm{s},\textbf{x}_\mathrm{g} \in \mathcal{Q}$.
However, this may result in the inclusion of "trivial" queries in the dataset, for which $\mathsf{steerTo}(\textbf{x}_\mathrm{s},\textbf{x}_\mathrm{g})=\mathsf{True}$. 
In case of such trivial queries, the neural planning Algorithm \ref{alg:neuralplanner} terminates in the first iteration after processing lines 4-6.
Thus, a key observation is that $\mathsf{steerTo}$ procedure in Algorithm \ref{alg:neuralplanner}, line 4 performs an implicit classification of queries, so that only "non-trivial" queries are passed over to the $\mathsf{PNet}$.
Concretely, the set of non-trivial queries can be defined as
\begin{equation}
\label{eq:nonTrivialQueries}
\mathcal{Q}_\mathrm{nt}=\{\textbf{x}_\mathrm{s},\textbf{x}_\mathrm{g} \in \mathcal{Q} \ | \ \mathsf{steerTo}(\textbf{x}_\mathrm{s},\textbf{x}_\mathrm{g})=\mathsf{False}   \}.
\end{equation}
This motivates the proposed data generation Algorithm \ref{alg:dataGen}, which aims to increase the number of non-trivial data samples in $\mathcal{D}$.
After an environment $\mathcal{X}, \mathcal{X}_\mathrm{obs} $ is created, data is generated by solving a total of $K_\mathrm{train}$ queries. 
With probability $p_\mathrm{nt}$, the proposed $\mathsf{nonTrivialQuerySampling}$ procedure is used to obtain $\textbf{x}_\mathrm{s},\textbf{x}_\mathrm{g} \in \mathcal{Q}_\mathrm{nt}$. 
Else, conventional $\mathsf{uniformSampling}$ returns a query in $\mathcal{Q}$.
A classical planner such as A* or BIT* then solves this query and outputs a good quality solution path $\pi$.
Samples from this path $\pi$ are appended to the dataset $\mathcal{D}$ with the proposed data inclusion procedure $\mathsf{includeData}$.

\subsection{Non-trivial Query Sampling}
A rejection sampling algorithm to generate new queries in $\mathcal{Q}_\mathrm{nt}$ is given in Algorithm \ref{alg:nonTrivialQuerySampling}.
For $N_\mathrm{nt}$ number of attempts, uniform sampling is used to first generate a valid query $\textbf{x}_\mathrm{s},\textbf{x}_\mathrm{g} \in \mathcal{Q}$. 
The $\mathsf{steerTo}$ module then validates the connection between start and goal state. 
If found invalid, the corresponding non-trivial query is returned.
Thus, this procedure intends to filter out trivial paths while maintaining the exploratory/coverage property of uniform sampling. 
Please see Fig. \ref{fig:point_robot_plot} and Fig. \ref{fig:nlink_plot} for a visualization of queries generated using the proposed non-trivial sampling procedure.

\subsection{Data Inclusion}
The data inclusion Algorithm \ref{alg:includeData} iterates over the segments of path $\pi$ and logs the current state, goal state $(\pi_{i},\pi_\mathrm{end})$ as the input and the next state $(\pi_{i+1})$ as the output/label.
However, if the flag $\mathsf{pruneData}=\mathsf{True}$, the algorithm skips including the data-sample $\{(\pi_{i},\pi_\mathrm{end}),\pi_{i+1} \}$ if $\pi_{i},\pi_\mathrm{end} \not\in \mathcal{Q}_\mathrm{nt}$. 
Thus, $\mathsf{pruneData}=\mathsf{True}$ ensures that only the non-trival segments of $\pi$ are incorporated in $\mathcal{D}$. 
Please see Fig. \ref{fig:non_trivial_schematic} for an illustration of this step.

Depending on the topology of $\mathcal{X}_\mathrm{obs}$, the neural planner may find it relatively harder to predict feasible paths in certain environments. 
The notion of non-trivial queries can be used to define a metric that captures this level of difficulty. 
Consider a "non-triviality ratio", which can be defined as,
\begin{equation}
    \gamma_\mathrm{nt}=\frac{\text{\# Non-trivial queries}}{\text{\# Uniformly sampled queries}} .
\end{equation}
Thus, $\gamma_\mathrm{nt}$ is the ratio of number of non-trivial queries found in a (large enough) set of uniformly sampled queries. 
This ratio will be high for complex, cluttered and narrow-passage type environments and low for relatively simpler, single-obstacle type environments.

\IncMargin{.5em}
\begin{algorithm}[t]
	\caption{Non-trivial Query Sampling}
	\label{alg:nonTrivialQuerySampling}
	\SetKwFunction{nonTrivialQuerySampling}{}
	\SetKwProg{Fn}{nonTrivialQuerySampling}{:}{}
	\Fn{\nonTrivialQuerySampling{ $\mathcal{X}_\mathrm{obs}$ }}
	{
	    \For{$i=1:N_\mathrm{max}$}
	    {
	        $(\textbf{x}_\mathrm{s},\textbf{x}_\mathrm{g}) \leftarrow \mathsf{uniformSampling}(\mathcal{X}_\mathrm{free})$\;
	        $\mathsf{valid} \leftarrow \mathsf{steerTo}(\textbf{x}_\mathrm{s},\textbf{x}_\mathrm{g})$\;
	        \If{$\mathsf{not ~ valid}$}
	        {
	            \KwRet $(\textbf{x}_\mathrm{s},\textbf{x}_\mathrm{g})$\;
	        }
	    }
	}
	\KwRet $(\textbf{x}_\mathrm{s},\textbf{x}_\mathrm{g})$\;
\end{algorithm}
\DecMargin{.5em}
\IncMargin{.5em}
\begin{algorithm}[t]
	\caption{Data Inclusion Procedure}
	\label{alg:includeData}
	\SetKwFunction{includeData}{}
	\SetKwProg{Fn}{includeData}{:}{}
	\Fn{\includeData{ $\mathcal{D},\pi$ }}
	{
	    \For{$i=1:L-1$}
	    {
	        \If{$\mathsf{pruneData}$}
	        {
	           $\mathsf{valid} \leftarrow \mathsf{steerTo}(\pi_{i},\pi_\mathrm{end})$\;
	           \If{$\mathsf{valid}$}
    	        {
    	            $\mathsf{continue}$\;
    	        }
	        }
	        $\mathcal{D} \leftarrow \mathcal{D} \cup \{(\pi_{i},\pi_\mathrm{end}),\pi_{i+1} \}$
	    }
	}
	\KwRet $\mathcal{D}$\;
\end{algorithm}
\DecMargin{.5em}

\section{NUMERICAL EXPERIMENTS}
In order to benchmark the proposed data generation algorithm, the following procedure was implemented for each planning environment.
First, four datasets with different parameter settings were created. 
These were as follows: $\mathcal{D}_0 ~( p_\mathrm{nt}=0, ~ \mathsf{pruneData}=\mathsf{False})$, $\mathcal{D}_1 ~( p_\mathrm{nt}=0.5,~ \mathsf{pruneData}=\mathsf{False})$, $\mathcal{D}_2 ~( p_\mathrm{nt}=1.0,~ \mathsf{pruneData}=\mathsf{False})$, $\mathcal{D}_3 ~( p_\mathrm{nt}=1.0,~ \mathsf{pruneData}=\mathsf{True})$.
Thus, $\mathcal{D}_0$  represents the dataset generated using the conventional uniform query sampling, whereas $\mathcal{D}_3$ is created using the proposed non-trivial query sampling and data pruning procedure. 
Four deep models, $\mathsf{PNet}_0,\mathsf{PNet}_1,\mathsf{PNet}_2,\mathsf{PNet}_3$ were then trained on their respective datasets.
The neural network shape, size and the training parameters were held constant while learning all four models.
Performance of the neural planner \ref{alg:neuralplanner} using these four models was evaluated on 1) $K_\mathrm{test}$ number of new uniform queries and 2) $K_\mathrm{test}$ number of new non-trivial queries.
Two performance metrics, namely, success ratio and cost ratio were considered. 
Success ratio gives the number of times (out of $K_\mathrm{test}$ in total) the neural planner was successful in finding a feasible (collision-free) solution.
Cost ratio denotes the ratio of the neural planner's solution cost to that of classical planner, averaged over $K_\mathrm{test}$ trials.  
Model training and evaluation was performed using the Python PyTorch API on a 64 bit, 16 GB RAM laptop with Intel i7 processor and a NVIDIA GeForce RTX 2060 GPU.
A description of robotic planning tasks along with a discussion of results is given below.

\begin{figure}
    \centering
    \includegraphics[width=0.42\columnwidth]{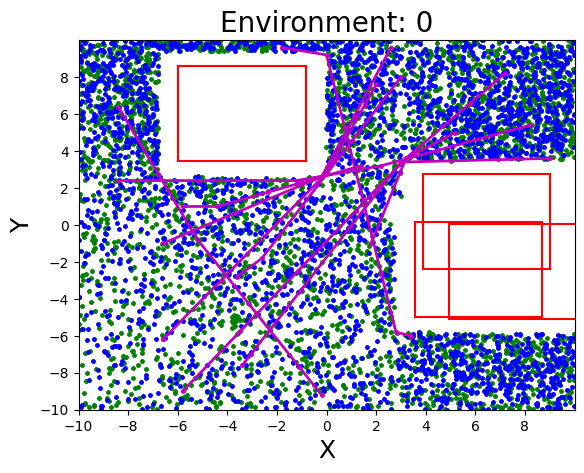}
	\includegraphics[width=0.42\columnwidth]{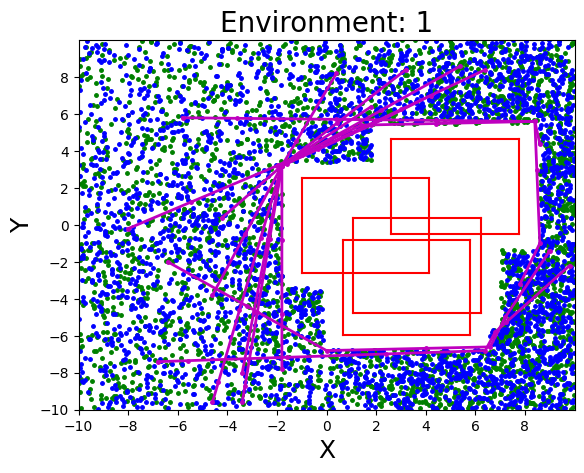}
	\includegraphics[width=0.42\columnwidth]{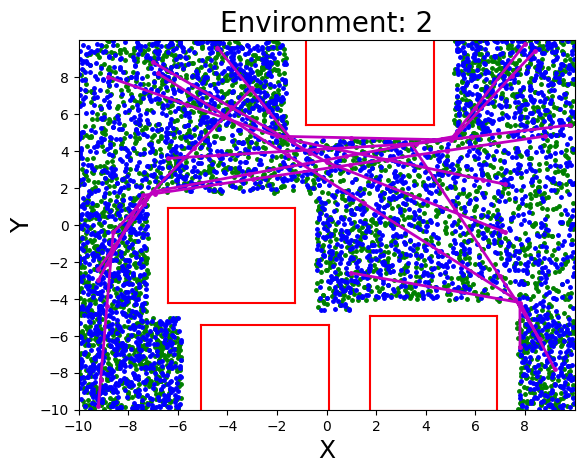}
	\includegraphics[width=0.42\columnwidth]{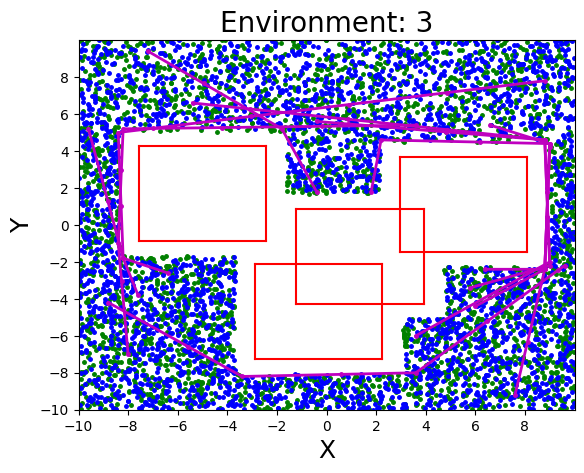}
	\includegraphics[width=.8\columnwidth,height=0.09\columnwidth]{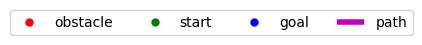}
	\caption{Four different environments for the point robot planning task. Data generated using the proposed non-trivial query sampling results in following (\textcolor{green}{start}, \textcolor{blue}{goal}) distribution. Some of the length-optimal paths solved using a classical planner are illustrated in \textcolor{magenta}{magenta}}
	\label{fig:point_robot_plot}
\end{figure}

\noindent \textbf{Point Robot:}
Four different $20 \times 20$ environments, illustrated in Fig. \ref{fig:point_robot_plot}, were considered for the case of point robot planning.
Four datasets $\{ \mathcal{D}_i \}^{3}_{i=0}$, as described above, were generated for each environment. 
A total of $K_\mathrm{train}=3000$ number of queries were sampled for each dataset.
An A* planner, followed by post-processing/smoothening, was used to solve these queries and obtain length-optimal paths.
A small padding of $0.8$ units around the obstacles was propagated during the data generation step, and was relaxed during the final performance evaluation step. 
This was found to greatly boost the success ratio of the neural planner, while making slight compromise in the cost ratio metric. 
The performance metrics were logged by solving $K_\mathrm{test}=500$ number of unseen uniform and non-trivial queries with the four learned $\mathsf{PNet}$ models.

\begin{figure}
    \centering
    \includegraphics[width=0.42\columnwidth]{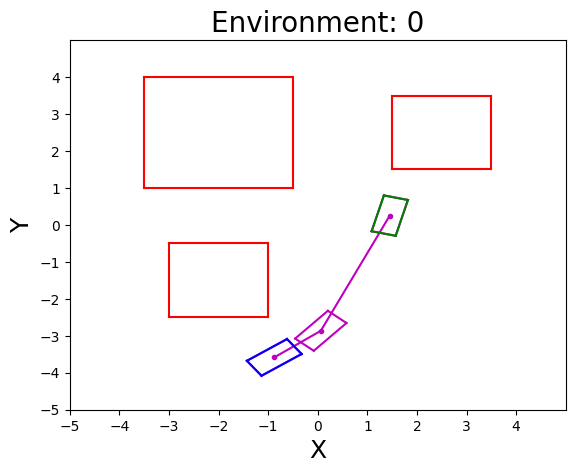}
	\includegraphics[width=0.42\columnwidth]{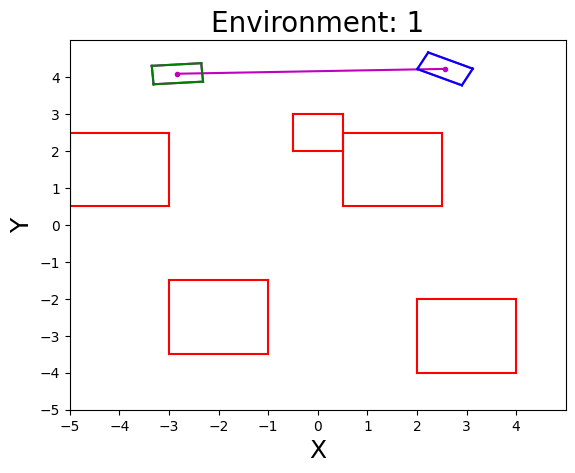}
	\includegraphics[width=0.42\columnwidth]{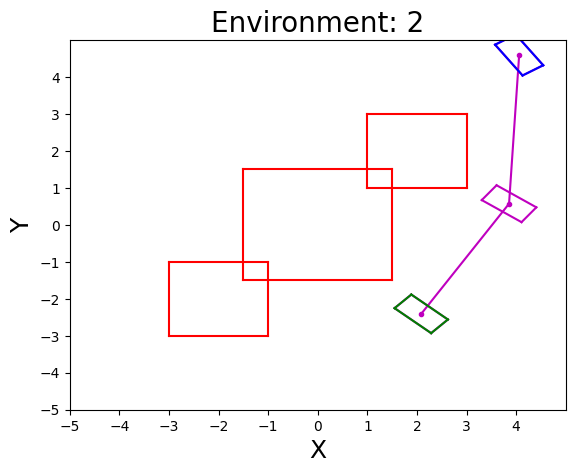}
	\includegraphics[width=0.42\columnwidth]{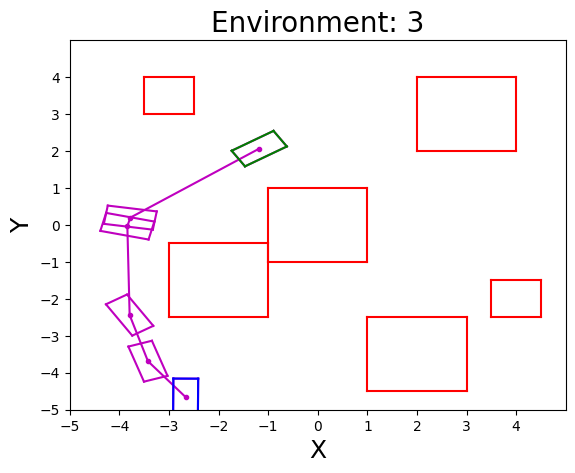}
	\includegraphics[width=.8\columnwidth,height=0.09\columnwidth]{figures/legend_nlink.png}
	\caption{Four different environments for the rigid body planning task.}
	\label{fig:rigid_body_plot}
\end{figure}

\noindent \textbf{Rigid Body Planning:}
Fig. \ref{fig:rigid_body_plot} shows the instance of planning for a rigid robot in four $10 \times 10$ environments. 
A total of $K_\mathrm{train}=5000$ queries were considered to create each of the four datasets $\{ \mathcal{D}_i \}^{3}_{i=0}$.
All the queries were solved in the $\mathrm{SE}(2)$ space using OMPL's \cite{sucan2012open} implementation of the BIT* planner.
An obstacle-padding of $0.4$ units was propagated during the data-generation phase. The learned $\mathsf{PNet}$ models predicted a three dimensional $[x,y,\theta]$ vector representing the robot's pose. 
These models were evaluated on $K_\mathrm{test}=500$ unseen uniform and non-trivial queries. 
The BIT* planner was allowed a run-time of $3$ seconds during the data-generation phase and $2$ seconds during the evaluation phase.

\begin{figure*}
	\centering
	\includegraphics[width=0.6\columnwidth]{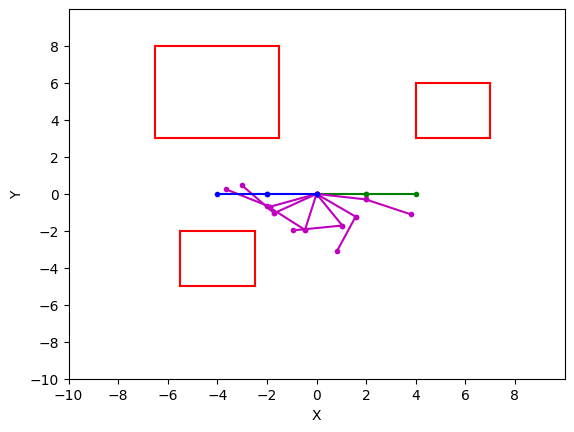}
	\includegraphics[width=0.6\columnwidth]{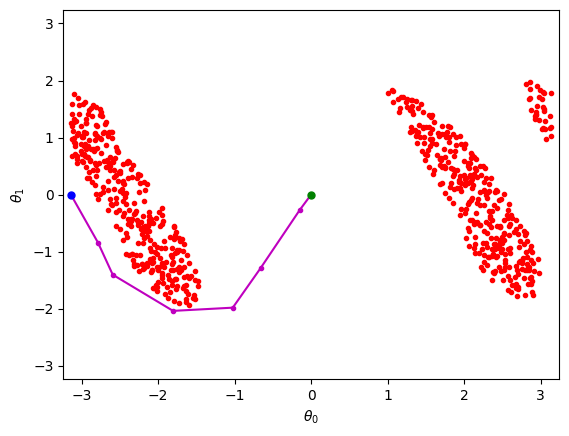}
	\includegraphics[width=0.6\columnwidth]{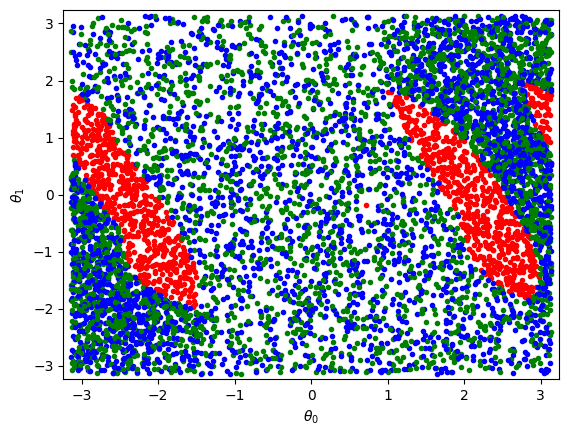}
	\includegraphics[width=.8\columnwidth,height=0.09\columnwidth]{figures/legend_nlink.png}
	\caption{A schematic for $n$-link manipulator planning. Left figure shows the obstacles and solution path in the work-space of the robot. The corresponding obstacles and solution path in the configuration-space (middle). The (\textcolor{green}{start}, \textcolor{blue}{goal}) distribution produced by the proposed non-trivial query sampling in the configuration-space (right).}
	\label{fig:nlink_plot}
\end{figure*}

\noindent \textbf{$n$-link Manipulator Planning:}
To observe performance of the neural planner in higher dimensions, a planning problem for $2,4$ and $6$-link manipulator robot was considered. Please see Fig. \ref{fig:nlink_plot}.
The joint angles were constrained to lie between $-\pi$ and $\pi$.
Four datasets $\{ \mathcal{D}_i \}^{3}_{i=0}$ were created for each of the $2,4$ and $6$-link case by considering a total of $3000,4000$ and $5000$ queries respectively. 
These queries were solved using OMPL's BIT* planner with an padding of $0.8$ units around the workspace obstacles.
The final performance evaluation was done by solving $K_\mathrm{test}=500$ new queries with the neural planner.
The BIT* planner was run for $2,4,6$ seconds during the data-generation stage and $1,2,4$ seconds during the evaluation stage for the $2,4$ and $6$-link planning respectively.

\begin{table}
	\caption{Point robot planning}
	\label{tab:point_robot_results}
	\centering
	Environment $0$, $\gamma_\mathrm{nt}=0.311$ \\
	\vspace{0.2em}
	\begin{tabular}{lSSSSSS}
		\toprule
		\multirow{2}{*}{Model} &
		\multicolumn{2}{c}{Uniform Query} &
		\multicolumn{2}{c}{Non-trivial Query} \\
		& {success ratio} & {cost ratio} & {success ratio} & {cost ratio} \\
		\midrule
		$\mathsf{PNet}_0$ & 0.970 & 1.004 & 0.894 & 1.006 \\
		$\mathsf{PNet}_1$ & 0.980 & 1.002 & 0.954 & 1.007 \\
		$\mathsf{PNet}_2$ & 0.974 & 1.003 & 0.918 & 1.010 \\
		$\mathsf{PNet}_3$ & 0.980 & 1.002 & 0.930 & 1.007  \\
		\bottomrule
	\end{tabular}
	\vspace{0.5em}
	\\Environment $1$, $\gamma_\mathrm{nt}=0.402$ \\
	\vspace{0.2em}
	\begin{tabular}{lSSSSSS}
		\toprule
		\multirow{2}{*}{Model} &
		\multicolumn{2}{c}{Uniform Query} &
		\multicolumn{2}{c}{Non-trivial Query} \\
		& {success ratio} & {cost ratio} & {success ratio} & {cost ratio} \\
		\midrule
		$\mathsf{PNet}_0$ & 0.992 & 1.031 & 0.986 & 1.075 \\
		$\mathsf{PNet}_1$ & 0.976 & 1.029 & 0.940 & 1.064 \\
		$\mathsf{PNet}_2$ & 0.986 & 1.027 & 0.966 & 1.063 \\
		$\mathsf{PNet}_3$ & 0.992 & 1.029 & 0.988 & 1.067 \\
		\bottomrule
	\end{tabular}
	\vspace{0.5em}
	\\Environment $2$, $\gamma_\mathrm{nt}=0.414$ \\
	\vspace{0.2em}
	\begin{tabular}{lSSSSSS}
		\toprule
		\multirow{2}{*}{Model} &
		\multicolumn{2}{c}{Uniform Query} &
		\multicolumn{2}{c}{Non-trivial Query} \\
		& {success ratio} & {cost ratio} & {success ratio} & {cost ratio} \\
		\midrule
		$\mathsf{PNet}_0$ & 0.966 & 1.051 & 0.880 & 1.117 \\
		$\mathsf{PNet}_1$ & 0.960 & 1.051 & 0.906 & 1.125 \\
		$\mathsf{PNet}_2$ & 0.950 & 1.044 & 0.916 & 1.120 \\
		$\mathsf{PNet}_3$ & 0.962 & 1.056 & 0.900 & 1.113 \\
		\bottomrule
	\end{tabular}
	\vspace{0.5em}
	\\Environment $3$, $\gamma_\mathrm{nt}=0.628$ \\
	\vspace{0.2em}
	\begin{tabular}{lSSSSSS}
		\toprule
		\multirow{2}{*}{Model} &
		\multicolumn{2}{c}{Uniform Query} &
		\multicolumn{2}{c}{Non-trivial Query} \\
		& {success ratio} & {cost ratio} & {success ratio} & {cost ratio} \\
		\midrule
		$\mathsf{PNet}_0$ & 0.944 & 1.056 & 0.864 & 1.156 \\
		$\mathsf{PNet}_1$ & 0.958 & 1.069 & 0.896 & 1.167 \\
		$\mathsf{PNet}_2$ & 0.956 & 1.063 & 0.898 & 1.163 \\
		$\mathsf{PNet}_3$ & 0.974 & 1.095 & 0.912 & 1.191 \\
		\bottomrule
	\end{tabular}
\end{table}

\begin{table}
	\caption{Rigid body planning}
	\label{tab:rigid_body_results}
	\centering
	Environment $0$, $\gamma_\mathrm{nt}=0.503$ \\
	\vspace{0.2em}
	\begin{tabular}{lSSSSSS}
		\toprule
		\multirow{2}{*}{Model} &
		\multicolumn{2}{c}{Uniform Query} &
		\multicolumn{2}{c}{Non-trivial Query} \\
		& {success ratio} & {cost ratio} & {success ratio} & {cost ratio} \\
		\midrule
		$\mathsf{PNet}_0$ & 0.828 & 0.985 & 0.656 & 0.972 \\
		$\mathsf{PNet}_1$ & 0.842 & 0.985 & 0.622 & 0.972 \\
		$\mathsf{PNet}_2$ & 0.834 & 0.979 & 0.698 & 0.960 \\
		$\mathsf{PNet}_3$ & 0.856 & 0.978 & 0.784 & 0.958  \\
		\bottomrule
	\end{tabular}
	\vspace{0.5em}
	\\Environment $1$, $\gamma_\mathrm{nt}=0.659$ \\
	\vspace{0.2em}
	\begin{tabular}{lSSSSSS}
		\toprule
		\multirow{2}{*}{Model} &
		\multicolumn{2}{c}{Uniform Query} &
		\multicolumn{2}{c}{Non-trivial Query} \\
		& {success ratio} & {cost ratio} & {success ratio} & {cost ratio} \\
		\midrule
		$\mathsf{PNet}_0$ & 0.788 & 0.948 & 0.714 & 0.874 \\
		$\mathsf{PNet}_1$ & 0.784 & 0.948 & 0.686 & 0.878 \\
		$\mathsf{PNet}_2$ & 0.792 & 0.971 & 0.638 & 0.889 \\
		$\mathsf{PNet}_3$ & 0.856 & 0.950 & 0.776 & 0.865 \\
		\bottomrule
	\end{tabular}
	\vspace{0.5em}
	\\Environment $2$, $\gamma_\mathrm{nt}=0.556$ \\
	\vspace{0.2em}
	\begin{tabular}{lSSSSSS}
		\toprule
		\multirow{2}{*}{Model} &
		\multicolumn{2}{c}{Uniform Query} &
		\multicolumn{2}{c}{Non-trivial Query} \\
		& {success ratio} & {cost ratio} & {success ratio} & {cost ratio} \\
		\midrule
		$\mathsf{PNet}_0$ & 0.862 & 0.965 & 0.770 & 0.900 \\
		$\mathsf{PNet}_1$ & 0.898 & 0.963 & 0.776 & 0.907 \\
		$\mathsf{PNet}_2$ & 0.902 & 0.963 & 0.806 & 0.908 \\
		$\mathsf{PNet}_3$ & 0.922 & 0.970 & 0.876 & 0.912 \\
		\bottomrule
	\end{tabular}
	\vspace{0.5em}
	\\Environment $3$, $\gamma_\mathrm{nt}=0.690$ \\
	\vspace{0.2em}
	\begin{tabular}{lSSSSSS}
		\toprule
		\multirow{2}{*}{Model} &
		\multicolumn{2}{c}{Uniform Query} &
		\multicolumn{2}{c}{Non-trivial Query} \\
		& {success ratio} & {cost ratio} & {success ratio} & {cost ratio} \\
		\midrule
		$\mathsf{PNet}_0$ & 0.722 & 0.985 & 0.606 & 1.000 \\
		$\mathsf{PNet}_1$ & 0.714 & 0.991 & 0.660 & 1.042 \\
		$\mathsf{PNet}_2$ & 0.756 & 0.994 & 0.650 & 0.996 \\
		$\mathsf{PNet}_3$ & 0.772 & 0.981 & 0.652 & 1.006 \\
		\bottomrule
	\end{tabular}
\end{table}

General trends seen from the results in Table \ref{tab:point_robot_results}, \ref{tab:rigid_body_results}, \ref{tab:n_link_results} are as follows. 
In most cases, $\mathsf{PNet}_2$ and $\mathsf{PNet}_3$ outperform $\mathsf{PNet}_0$ in terms of success ratio, where as the performance of $\mathsf{PNet}_1$ is more sporadic. 
The cost ratio is relatively lower for the rigid body and $6$-link case compared to others, as the BIT* planner may not find a good quality solution in the given planning time for these challenging cases. 
The success ratio for all $\mathsf{PNet}$ models is naturally higher over uniform test queries rather than non-trivial test queries.
The success ratio also generally has an inverse relation with $\gamma_\mathrm{nt}$, as seen strongly in the case of rigid body and $n$-link manipulator planning.
For the case of point robot planning, all models perform well with a success rate of over $90 \%$ (see Table \ref{tab:point_robot_results}). 
However, slight performance gains due to the proposed method can be seen for Environments $0,2,3$. 
These gains are much more noticeable for the rigid body planning case (see Table \ref{tab:rigid_body_results}). 
The $\mathsf{PNet}_3$ model has the highest success ratio in all cases except one (Environment $3$, Non-trivial Query), where its performance is comparable to $\mathsf{PNet}_1$. 
The gradation in performance due to dimensionality and $\gamma_\mathrm{nt}$ can be seen clearly in the $n$-link planning case (see Table \ref{tab:n_link_results}). 
The success ratio over uniform queries is in the range of $0.9,0.8$ and $0.7$ for the case of $2,4$ and $6$-link planning case respectively.
For the relatively simpler $2$-link planning case with $\gamma_\mathrm{nt}=0.225$, only small gains in the success ratio over non-trivial queries can be seen. 
However, the improvement in performance is much more evident for the higher dimensional $4$ and $6$-link cases.  
The $\mathsf{PNet}_3$ model shows about a $25\%$ increase in the success ratio over $\mathsf{PNet}_0$ for the $6$-link (non-trivial queries) case.

\begin{figure}
    \centering
    \includegraphics[width=0.41\columnwidth]{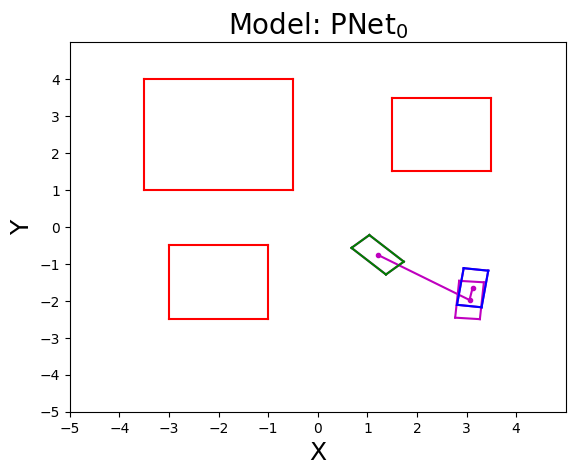}
	\includegraphics[width=0.41\columnwidth]{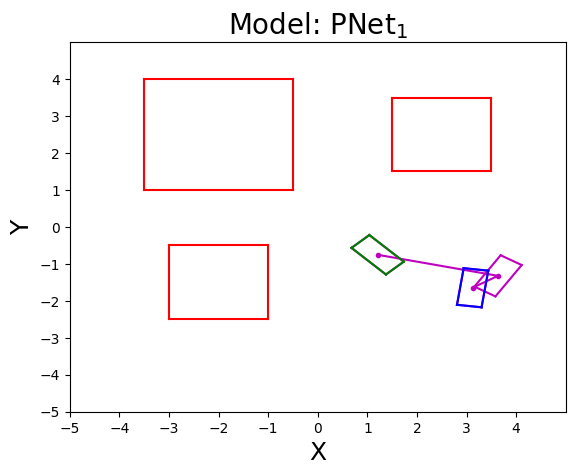}
	\includegraphics[width=0.41\columnwidth]{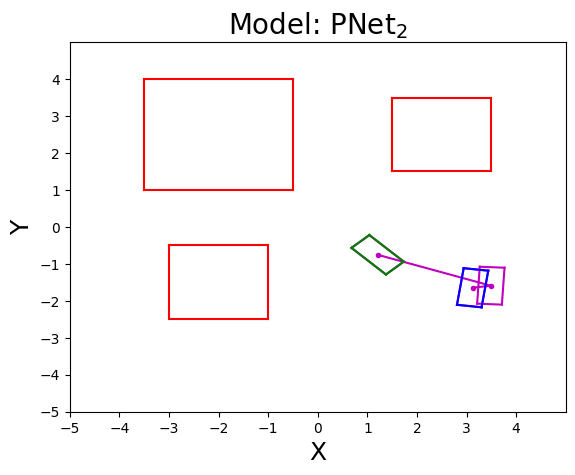}
	\includegraphics[width=0.41\columnwidth]{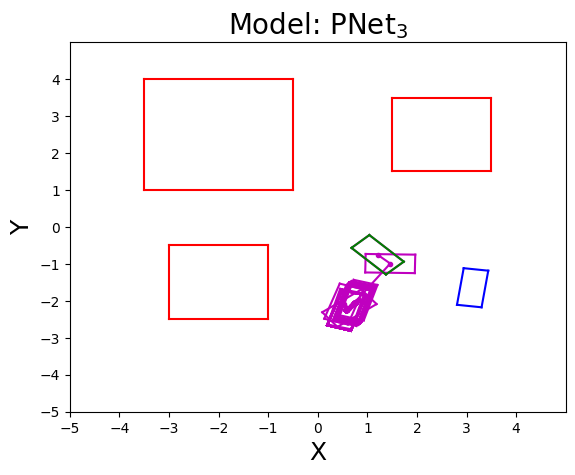}
	\includegraphics[width=.8\columnwidth,height=0.08\columnwidth]{figures/legend_nlink.png}
	\caption{Solving a trivial query with the four learned models without the $\mathsf{steerTo}$ function. Models $\mathsf{PNet}_0$,$\mathsf{PNet}_1$,$\mathsf{PNet}_2$ can successfully solve the query. However, the $\mathsf{PNet}_3$ model, which does not have any trivial sample in its training dataset, is unable to solve it.}
	\label{fig:rigid_body_pnet3_fail}
\end{figure}

The neural planning Algorithm \ref{alg:neuralplanner} and the corresponding results discussed above assume the availability of a steering function. 
While this is readily available for cases such as geometric or non-holonomic (car-like) planning \cite{johnson2020dynamically}, it may not be computationally tractable for others. 
To analyze the performance of $\mathsf{PNet}$ models without the $\mathsf{steerTo}$ function, simulations were performed by only executing the lines $8$ and $9$ of the neural planner \ref{alg:neuralplanner} for maximum $N_\mathrm{plan}$ iterations. 
Instead of lines 4-6 in Algorithm \ref{alg:neuralplanner}, the following termination condition was implemented, $\| \pi_\mathrm{end} - \textbf{x}_\mathrm{g} \|_2 \leq \delta$, with a small $\delta>0$.
As illustrated in Fig. \ref{fig:rigid_body_pnet3_fail}, the $\mathsf{PNet}_3$ model, which has no trivial sample in its training dataset, naturally cannot solve a trivial query. 
Numerical results for the rigid body planning without the $\mathsf{steerTo}$ function and $\delta=1.0$ are tabulated in Table \ref{tab:rigid_body_nosteer_results}.
The success ratio of all $\mathsf{PNet}$ models is adversely affected in this case. 
The $\mathsf{PNet}_0$ model performs best on uniform queries in all environments, whereas the performance of $\mathsf{PNet}_3$ is the worst. 
However, $\mathsf{PNet}_1$ or $\mathsf{PNet}_2$ show better performance over non-trivial queries in some cases. 
Thus, without a steering function, a uniformly sampled training dataset might be the best choice if the test queries are uniformly distributed too. 
However, a model trained over a dataset with an appropriate value of $p_\mathrm{nt}$ may perform better over non-trivial test queries. 
This makes a case for an ensemble model.

\begin{table}
	\caption{Rigid Body without $\mathsf{steerTo}$ }
	\label{tab:rigid_body_nosteer_results}
	\centering
	Environment $0$, $\gamma_\mathrm{nt}=0.503$ \\
	\vspace{0.2em}
	\begin{tabular}{lSSSSSS}
		\toprule
		\multirow{2}{*}{Model} &
		\multicolumn{2}{c}{Uniform Query} &
		\multicolumn{2}{c}{Non-trivial Query} \\
		& {success ratio} & {cost ratio} & {success ratio} & {cost ratio} \\
		\midrule
		$\mathsf{PNet}_0$ & 0.624 & 0.982 & 0.414 & 0.957 \\
		$\mathsf{PNet}_1$ & 0.592 & 0.983 & 0.372 & 0.979 \\
		$\mathsf{PNet}_2$ & 0.564 & 0.977 & 0.392 & 0.960 \\
		$\mathsf{PNet}_3$ & 0.228 & 0.985 & 0.212 & 0.978  \\
		\bottomrule
	\end{tabular}
	\vspace{0.5em}
	\\Environment $1$, $\gamma_\mathrm{nt}=0.659$ \\
	\vspace{0.2em}
	\begin{tabular}{lSSSSSS}
		\toprule
		\multirow{2}{*}{Model} &
		\multicolumn{2}{c}{Uniform Query} &
		\multicolumn{2}{c}{Non-trivial Query} \\
		& {success ratio} & {cost ratio} & {success ratio} & {cost ratio} \\
		\midrule
		$\mathsf{PNet}_0$ & 0.498 & 0.971 & 0.380 & 0.936 \\
		$\mathsf{PNet}_1$ & 0.476 & 0.965 & 0.430 & 0.923 \\
		$\mathsf{PNet}_2$ & 0.456 & 0.974 & 0.372 & 0.946 \\
		$\mathsf{PNet}_3$ & 0.266 & 0.970 & 0.208 & 0.968 \\
		\bottomrule
	\end{tabular}
	\vspace{0.5em}
	\\Environment $2$, $\gamma_\mathrm{nt}=0.556$ \\
	\vspace{0.2em}
	\begin{tabular}{lSSSSSS}
		\toprule
		\multirow{2}{*}{Model} &
		\multicolumn{2}{c}{Uniform Query} &
		\multicolumn{2}{c}{Non-trivial Query} \\
		& {success ratio} & {cost ratio} & {success ratio} & {cost ratio} \\
		\midrule
		$\mathsf{PNet}_0$ & 0.684 & 0.943 & 0.514 & 0.872 \\
		$\mathsf{PNet}_1$ & 0.684 & 0.944 & 0.506 & 0.889 \\
		$\mathsf{PNet}_2$ & 0.630 & 0.929 & 0.550 & 0.890 \\
		$\mathsf{PNet}_3$ & 0.284 & 0.937 & 0.270 & 0.931 \\
		\bottomrule
	\end{tabular}
	\vspace{0.5em}
	\\Environment $3$, $\gamma_\mathrm{nt}=0.690$ \\
	\vspace{0.2em}
	\begin{tabular}{lSSSSSS}
		\toprule
		\multirow{2}{*}{Model} &
		\multicolumn{2}{c}{Uniform Query} &
		\multicolumn{2}{c}{Non-trivial Query} \\
		& {success ratio} & {cost ratio} & {success ratio} & {cost ratio} \\
		\midrule
		$\mathsf{PNet}_0$ & 0.514 & 0.986 & 0.374 & 0.990 \\
		$\mathsf{PNet}_1$ & 0.504 & 0.996 & 0.426 & 1.005 \\
		$\mathsf{PNet}_2$ & 0.502 & 1.004 & 0.396 & 1.004 \\
		$\mathsf{PNet}_3$ & 0.302 & 1.001 & 0.196 & 0.979 \\
		\bottomrule
	\end{tabular}
\end{table}
\begin{table}
	\caption{$n$-link manipulator planning}
	\label{tab:n_link_results}
	\centering
	$2$-link planning, $\gamma_\mathrm{nt}=0.225$ \\
	\vspace{0.2em}
	\begin{tabular}{lSSSSSS}
		\toprule
		\multirow{2}{*}{Model} &
		\multicolumn{2}{c}{Uniform Query} &
		\multicolumn{2}{c}{Non-trivial Query} \\
		& {success ratio} & {cost ratio} & {success ratio} & {cost ratio} \\
		\midrule
		$\mathsf{PNet}_0$ & 0.984 & 1.002 & 0.924 & 1.061 \\
		$\mathsf{PNet}_1$ & 0.990 & 1.004 & 0.956 & 1.060\\
		$\mathsf{PNet}_2$ & 0.984 & 1.003 & 0.962 & 1.062 \\
		$\mathsf{PNet}_3$ & 0.988 & 1.004 & 0.972 & 1.068 \\
		\bottomrule
	\end{tabular}
	\vspace{0.5em}
	\\ $4$-link planning, $\gamma_\mathrm{nt}=0.366$ \\
	\vspace{0.2em}
	\begin{tabular}{lSSSSSS}
		\toprule
		\multirow{2}{*}{Model} &
		\multicolumn{2}{c}{Uniform Query} &
		\multicolumn{2}{c}{Non-trivial Query} \\
		& {success ratio} & {cost ratio} & {success ratio} & {cost ratio} \\
		\midrule
		$\mathsf{PNet}_0$ & 0.862 & 1.005 & 0.722 & 1.003 \\
		$\mathsf{PNet}_1$ & 0.876 & 0.997 & 0.718 & 0.991\\
		$\mathsf{PNet}_2$ & 0.890 & 0.994 & 0.772 & 1.003 \\
		$\mathsf{PNet}_3$ & 0.928 & 0.860 & 0.82 & 0.992 \\
		\bottomrule
	\end{tabular}
	\vspace{0.5em}
	\\ $6$-link planning, $\gamma_\mathrm{nt}=0.608$ \\
	\vspace{0.2em}
	\begin{tabular}{lSSSSSS}
		\toprule
		\multirow{2}{*}{Model} &
		\multicolumn{2}{c}{Uniform Query} &
		\multicolumn{2}{c}{Non-trivial Query} \\
		& {success ratio} & {cost ratio} & {success ratio} & {cost ratio} \\
		\midrule
		$\mathsf{PNet}_0$ & 0.716 & 0.982 & 0.512 & 0.972 \\
		$\mathsf{PNet}_1$ & 0.650 & 0.980 & 0.454 & 0.966\\
		$\mathsf{PNet}_2$ & 0.712 & 0.984 & 0.528 & 0.971 \\
		$\mathsf{PNet}_3$ & 0.776 & 0.975 & 0.644 & 0.958 \\
		\bottomrule
	\end{tabular}
\end{table}

\section{CONCLUSION}
Many of the previous techniques in the literature have focused on exploring different deep architectures for planning, while using a uniformly sampled dataset for training.  
This work, on the other hand, investigates the problem of improving the data-generation process while holding the model architecture and planning algorithm constant.
The proposed query sampling and data pruning procedures add more complicated paths in the dataset. 
Numerical experiments show that the success rate of the neural planner can be boosted using the deep models trained on such non-trivial datasets.

This work presents many opportunities for future research.
An ensemble model can be constructed by combining predictions from different models trained on datasets with varying degrees of non-triviality. 
Instead of a Boolean $\mathsf{pruneData}$ flag, calling the pruning procedure with a probability of $\gamma_\mathrm{nt}$ can be explored. 
This can prevent excessive pruning and result in a drastic reduction in the size of the dataset for relatively less cluttered environments. 

\noindent \textbf{Acknowledgements:} Authors would like to thank Prof. Le Song, Binghong Chen and Ethan Wang for insightful discussions on this topic.

\bibliographystyle{IEEEtran}
\bibliography{references}	

\end{document}